\documentclass[final,onefignum,onetabnum]{siuro210301}
\usepackage{tikz}
\usepackage{comment}
\usepackage{lineno}

\usepackage{float}
\usepackage{subcaption}
\usepackage{lipsum}
\usepackage{amsfonts}
\usepackage{graphicx}
\usepackage{epstopdf}
\usepackage{algorithmic}
\ifpdf
  \DeclareGraphicsExtensions{.eps,.pdf,.png,.jpg}
\else
  \DeclareGraphicsExtensions{.eps}
\fi

\newsiamremark{remark}{Remark}
\newsiamremark{hypothesis}{Hypothesis}
\crefname{hypothesis}{Hypothesis}{Hypotheses}
\newsiamthm{claim}{Claim}

\headers{}{G. Barone, A. Cunchala, and R. Nunez}

\title{Accurately Classifying Out-Of-Distribution Data in Facial Recognition\thanks{Submitted to the editors June, 2024.
\funding{This work was funded in part by the US NSF award DMS-2051019.}}}

\author{Gianluca Barone\thanks{Department of Mathematics, Rowan University, Glassboro, NJ 
  (\email{barone56@students.rowan.edu}).}
\and Aashrit Cunchala\thanks{Department of Applied Mathematics, University of Pittsburgh, Pittsburgh, PA 
  (\email{aac130@pitt.edu})}
\and Rudy Nunez\thanks{Department of Mathematics, Emory University, Atlanta, GA (\email{rudy.nunez@emory.edu})}}
\dedication{\small\textit{Project advisor: Nicole Yang}\thanks{Department of Mathematics, Emory University, Atlanta, GA (\email{tianjiao.yang@emory.edu})}}

\usepackage{amsopn}

\newcommand{\E}{\mathbb{E}}

\newcommand{\triplenorm}[1]{{\vert\kern-0.25ex\vert\kern-0.25ex\vert #1 
    \vert\kern-0.25ex\vert\kern-0.25ex\vert}}

\newtheorem{exm}{Example}[section]

\theoremstyle{mydefinition}
\theoremstyle{myremark}

\newcommand{\xdownarrow}[1]{%
  {\left\downarrow\vbox to #1{}\right.\kern-\nulldelimiterspace}
}
\ifpdf
\hypersetup{
  pdftitle={Accurately Classifying Out-Of-Distribution Data in Facial Recognition},
  pdfauthor={G. Barone, A. Cunchala, and R. Nunez }
}

\headers{Accurately Classifying Out-Of-Distribution Data in Facial Recognition}{G. Barone, A. Cunchala, and R. Nunez}

\begin{document}

\maketitle

\begin{abstract}
Standard classification theory assumes that the distribution of images in the test and training sets are identical. Unfortunately, real-life scenarios typically feature unseen data (``out-of-distribution data") which is different from data in the training distribution (``in-distribution"). This issue is most prevalent in social justice problems where data from under-represented groups may appear in the test data without representing an equal proportion of the training data. This may result in a model returning confidently wrong decisions and predictions. We are interested in the following question: Can the performance of a neural network improve on facial images of out-of-distribution data when it is trained simultaneously on multiple datasets of in-distribution data? We approach this problem by incorporating the Outlier Exposure model and investigate how the model's performance changes when other datasets of facial images were implemented. We observe that the accuracy and other metrics of the model can be increased by applying Outlier Exposure, incorporating a trainable weight parameter to increase the machine's emphasis on outlier images, and by re-weighting the importance of different class labels. We also experimented with whether sorting the images and determining outliers via image features would have more of an effect on the metrics than sorting by average pixel value, and found no conclusive results. Our goal was to make models not only more accurate but also more fair by scanning a more expanded range of images. Utilizing Python and the Pytorch package, we found models utilizing outlier exposure could result in more fair classification. 
\end{abstract}



\section{Introduction} 
Facial recognition is being used increasingly in the medical, security, and criminal investigation fields, along with many more as time passes. In an attempt to create machine learning models for classifying faces through facial recognition, many large data sets have been collected. However, in many of these data sets, there is a heavy representation of Caucasian faces, while other races are underrepresented. As a result, studies show that the skew in data sets can be around 80$\%$ in favor of lighter skin tones \cite{merler2019diversity}. This leads to situations where models fail to detect darker skin tones. One such example is when Microsoft, IBM, and Face++ classifiers were tested against data sets where it was shown they all perform best on lighter faces, and perform the worst on darker female faces \cite{buolamwini2018gender}. To combat this issue, some data sets have been curated to provide a more balanced division of groups. One such example is FairFace \cite{kärkkäinen2019fairface}, which was developed specifically to provide an equal division of race defined by 7 categories. Additionally, many problems that apply to race classification extend to gender classification, with some frequently used datasets containing a heavily unequal gender distribution \cite{merler2019diversity}.

The topic of reducing bias to create a more fair classification model has been widely studied in recent years, beyond just creating balanced data sets. Particularly, a model can struggle when encountering data with different distributions than the one present in the training data. In real-world applications, these distributional differences often create issues with the model's ability to detect minority groups due to the overpowering amount of majority group data. One approach to this problem is to instead expose the model to out-of-distribution (OOD) data during the training process through the use of Outlier Exposure \cite{hendrycks2018deep}. One potential shortcoming of this method is that in real life, information about the OOD data is not guaranteed. 

The contribution of our paper is
\begin{itemize}
    \item We compare the advantages between loss re-weighting and Outlier Exposure motivated by their different restrictions on the support of input and output data, and find that each increase different metrics of fairness. 
    \item We modify the Outlier Exposure accordingly in terms of the loss computation and the exposure parameter using the distance between the training and outlier distributions and observe that we can achieve better results through this method.
    \item We observe the effects of editing the weighting parameter on accuracy and observe that our changes result in fewer data points being falsely identified.
\end{itemize}

\section{Distribution Shifts in Classification}

Our focus is on the classification problem in facial recognition. Consider a large-scale facial image dataset containing N images, $\{\mathbf{x}_i, y_i\}_{i=1}^N$, where $\mathbf{x}_i \in \mathcal{X}$ denotes the $i$-th facial image and $y_i \in \{1, \ldots, K\}$ denotes the corresponding label, for example, the identity of the person in the image. The goal of a model is to approximate a function, $f(x)$, to relate the images to their labels based on the training samples $\{\mathbf{x}_i^{\text{train}}, y_i^{\text{train}}\}_{i=1}^n$. If this relationship is successfully established, then the labels $y^{\text{test}}$ can be found by $f(x^{\text{test}})$. In training, the input images $\mathbf{x}^{\text{train}}_1, \ldots, \mathbf{x}^{\text{train}}_n$ are independent and identically distributed random variables following the image distribution $P(\mathbf{x})$. The output labels of training data follow a conditional distribution given $\mathbf{x}^{\text{train}}$,
\[
 y_i^{\text{train}} \sim P(y | \mathbf{x} = \mathbf{x}_i^{\text{train}}).
\]
In standard classification problems, $x^{\text{train}}$ and $x^{\text{test}}$ are from identical distributions. However, in our application the two distributions are not the same. Instead, we have the training distribution $P(x^{\text{train}})$ and testing distribution $Q(x^{\text{test}})$.

\subsection{Major issues}
There are two major issues we encounter when working with this classification problem. The first one is data imbalance. The distributions of different classes in the two datasets are different. For example, FairFace, the training dataset, has a near 50/50 split between male and female faces while UTKFace, one of the testing datasets, is heavily slanted in favor of male faces. The other major issue we face is the two datasets being out-of-distribution. As we will show, the two datasets had non-overlapping parts which makes it difficult to classify images using typical classification strategies.

To correct the imbalances found within the training and test sets, we attempt to weight the sampling as well as weight the importance of each sample. Due to the fact that datasets always have an uneven amount of samples between classes, we strengthen the importance of the minority classes to help offset imbalances. As has been shown \cite{byrd2019effect}, given samples from a distribution $P(x)$, a target distribution $Q(x)$, and the function $f(x)$ approximating the samples under $Q(x)$, importance sampling would produce an unbiased estimate given by the likelihood ratio $\frac{Q(x)}{P(x)}$:

\begin{equation}
    \E_P\left[\frac{Q(x)}{P(x)}f(x)\right] = \E_Q[f(x)]
    \label{Degeneracy}
\end{equation}

For that reason, we focus on weighting the importance according to the likelihood ratio. We do this by manually tuning the re-scaling weight for each class based on the distribution of training data. The goal of class re-weighting is to show how the model's performance is affected by out-of distribution data instead of imbalanced distributions.

One of our main approaches is to implement Outlier Exposure using multiple datasets to accurately update parameters based on distributions. To do this, we generate an auxiliary dataset consisting of the outliers of a given dataset. We quantify the largest outliers through the use of the Kullback–Leibler (KL) divergence.

For each dataset, the images are sorted based on their KL distance from the distribution. We collect 20$\%$ of the data with the greatest KL distance from each dataset to create an outlier dataset containing the most significant outliers. We then use this new dataset for Outlier Exposure to expose the model to the most significant outliers. 

We find that re-weighting and Outlier Exposure both benefit performance, however in some cases one is superior to the other. We find that re-weighting balances the model's predictions for each class, whereas outlier exposure has a greater effect on improving accuracy and other metrics such as precision and recall. However, both techniques can be used simultaneously to complement each other, but it is important to consider how the weighted classes will affect the predictions made on the Outlier Exposure group. These weights were specified and the formula provided in Section 3.3.

\section{Methods}
\label{sec:main}
\subsection{Explaining Outlier Exposure} \label{sec:oe}
One common way to alleviate out-of-distribution data (OOD) missing their true labels is by using auxiliary out-of-distribution data. This auxiliary data can be incorporated into the model through an OOD-detection score that can then be used to generalize unseen data through the model \cite{hendrycks2018deep}. The loss function for this model is:

\begin{equation}
    \min_{\theta} \E_{P_{in}(\mathbf{x}, y)} [L(f_{\theta}(\mathbf{x}),y)] + \lambda \E_{P_{out}(\mathbf{x}^{OE})} [L(f_{\theta}(\mathbf{x}^{OE}), y^{OE)}],
    \label{outlier_exposure}
\end{equation}
where $x$ is an $1 \times n$ pixel-intensity vector of the in-distribution data, $y$ is the $1\times n$ true-labeling of the images contained in the in-distribution set, ${x}^{OE}$ is the $1\times n$ pixel-intensity vector of the auxiliary data-set, and ${y}^{OE}$ is a $1\times n$ vector of the true-labeling of the auxiliary dataset. The pixels are capped at $255$ due to the standard RGB scale. Every image in the dataset is a random variable $\mathbf{x} \in [0,1]^{n^2}$, where $n$ is the resolution value. The pixel average for the current image examined is defined as $\mathbf{x}_1$ and the overall distribution of pixel averages for all pixels is defined as $\mathbf{x}_2$, with the range for each of these averages between $0$ to $255$. 
The first term is a loss function designed to minimize the loss of the in-distribution data to increase model accuracy. The second term is a loss function, with separate predictions, designed to minimize the loss of out-of-distribution data in the model's decision making process. The parameter $\lambda$ is then trained to find the optimal ratio between the weights of the in-distribution and out-of distribution data during each epoch.

To clarify further,
\begin{equation}
    \E_{P_{in}(\mathbf{x}, y)} [L(f_{\theta}(\mathbf{x}),y) ]  
\end{equation}
is the loss for in-distribution data. It is a product between the expected value of a function of pixel-intensity and true-labeling and the cross-entropy loss function of the pixel-intensity and true-labeling.

$L(\cdot,\cdot)$ is the cross-entropy loss function, which measures the distance between the target probability distribution and the learned probability distribution. That is,
\begin{equation}
   \E_{P_{in}(\mathbf{x}, y)}  L(f_{\theta}(\mathbf{x}),y) = \E_{P_{in}(\mathbf{x}, y)}  [- \log P_{\theta}(\mathbf{x},y) ]. \label{OEIn}
\end{equation}

Since images can only belong to one class, the probability of the image belonging to the true label is 1 and so, the probability of the image belong to the other labels are 0. Therefore with $n$ samples, the expectation can be approximated by $\frac{1}{n} \sum_{i=1}^n L(f_{\theta}(\mathbf{x}_i),y_i)$. Therefore our equation changes to 

\begin{equation}
\begin{split}
   \E_{P_{in}(\mathbf{x}, y)}  L(f_{\theta}(\mathbf{x}),y) = \E_{P_{in}(\mathbf{x}, y)}  [- \log P_{\theta}(\mathbf{x},y) ], \\  \approx -\frac{1}{n} \sum_{j=1}^n \sum_{i = 1}^K \mathbf{1}[y_j = i] \log p_{\theta}(y_j = i| \mathbf{x}_j), 
\end{split}
\end{equation}
where the distribution $p_{\theta}(y_j = i| \mathbf{x}_j)$ is learned from a neural network by using the typical softmax function. The loss for Outlier Exposure data is

\begin{equation}
    \E_{P_{out}(\mathbf{x}^{OE})} [L(f_{\theta}(\mathbf{x}^{OE}), y^{OE}]. \label{OEOut}
\end{equation}
The Outlier Exposure we implement in our model is approximated by:
\begin{equation}
\begin{split}
\\  OE = -\frac{1}{n} \sum_{j=1}^n \sum_{i = 1}^K \mathbf{1}[y_j = i] \log p_{\theta}(y_j = i| \mathbf{x}_j) -\frac{1}{n} \sum_{j=1}^n \sum_{i = 1}^K \mathbf{1}[y_j = i] \log p_{\theta}(y_j = i| \mathbf{x}_j)], 
\label{NewOEOut}
\end{split}
\end{equation}
where $x_j$ is the in-distribution data set, UTKFace, $y_j$ are the true labeling, $x_{OE}$ is the out-of-distribution, and $y_{OE}$ is the true labeling of the out-of-distribution data where the out-of-distribution varies between different datasets. Essentially, $x_{OE}$ is the distribution of the model's class prediction for the test set and $y_{OE}$ is the true labeling of classes for the test set.

\subsection{Toy Example}
\begin{figure}[tph]
  \centering
  \includegraphics[width=0.5\textwidth]{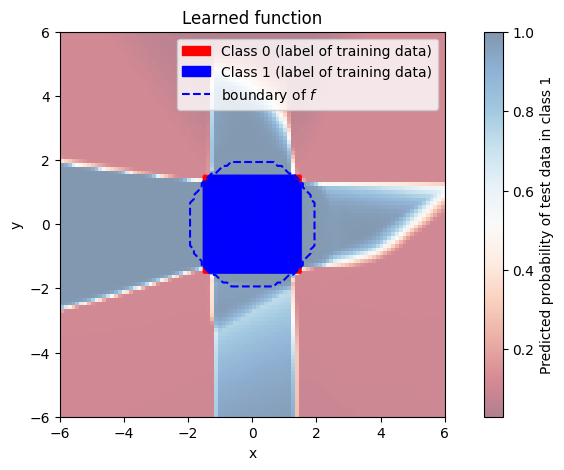}
  \caption{Any point inside the dashed circle gives $f(\mathbf{x}) = 1$ and its true label is class 1, while the true label of points in the rest of the whole space $[-6,6]^2$ is class 0. The training data $(\mathbf{x}^{\text{train}}, y)$ has inputs from the smaller square $\mathbf{x}^{\text{train}} \in [-1.5, 1.5]^2$, and $y = f(\mathbf{x}^{\text{train}})$. This can be seen in the middle small square with blue (class 1) in the overlapping part with $x_1^2 + x_2^2 \leq 4$; and red corners (class 0). The color bar on the side gives the predicted probability of the test data (a uniform mesh grid on $[-6,6]^2$) in class 1.}
  \label{fig:figure_label2}
\end{figure}

Let's consider two toy examples to visualize the difficulties a model encounters when dealing with OOD data. Let's make the classification of the model binary. We have the problem shown as below:
\begin{exm}
     Take input data $\mathbf{x} = (x_1,x_2)$, label $y = \{0,1 \}$. We first set up an explicit mapping $f: \mathbf{x} \rightarrow y$, where 
\begin{equation}
    \label{donut}
f(x_1,x_2) =
    \begin{cases}
      1 & \text{if $x_1^2 + x_2^2 \leq 4$}\\
      0 & \text{otherwise}.
    \end{cases}     
\end{equation}
The training data $(\mathbf{x}^{\text{train}}, y)$ has inputs from the smaller square $\mathbf{x}^{\text{train}} \in [-1.5, 1.5]^2$, and $y = f(\mathbf{x}^{\text{train}})$. The test data is a uniform mesh grid on $[-6,6]^2$. If the mapping and classification is correct all the area outside the circle should be red. We see that Fig.~\ref{fig:figure_label2} shows that the classification result is accurate for data points close to the given training data. However, the learned function fails to classify the cross-shaped area outside the circle correctly but is confident in making that incorrect prediction. This shows the learning algorithm over-confidently gives wrong classification results on the test data it has not seen before. In this example, we used a simple fully-connected neural network with 2 hidden layers and 100 neurons per layer and apply Adam optimizer to optimize the cross-entropy loss.  
\end{exm}

When the data provided is reliable (truly sampled from $P_{in}(\mathbf{x}, y)$) the classifier gives a good result. However, when encountering a different distribution, the model cannot classify accurately. We attempt to resolve this by introducing a form of adjustment for the model known as Outlier Exposure as described in Section~\ref{outlierexposure_section}, seen in the example below:
\begin{exm}
    The model has not seen data from $P_{out}$ during training and consequently that data may be incorrectly identified. As can be seen in Figure~\ref{fig:toy}, Outlier Exposure does allow for classification to be more accurate for different classes. The classifier is a simple case and can be defined explicitly as the donut shaped region. We use the same neural network setup as seen in the initial example.

\begin{figure}[tph]
    \centering
    \begin{subfigure}{0.75\textwidth}
    \begin{center}
        \includegraphics[scale=0.25]{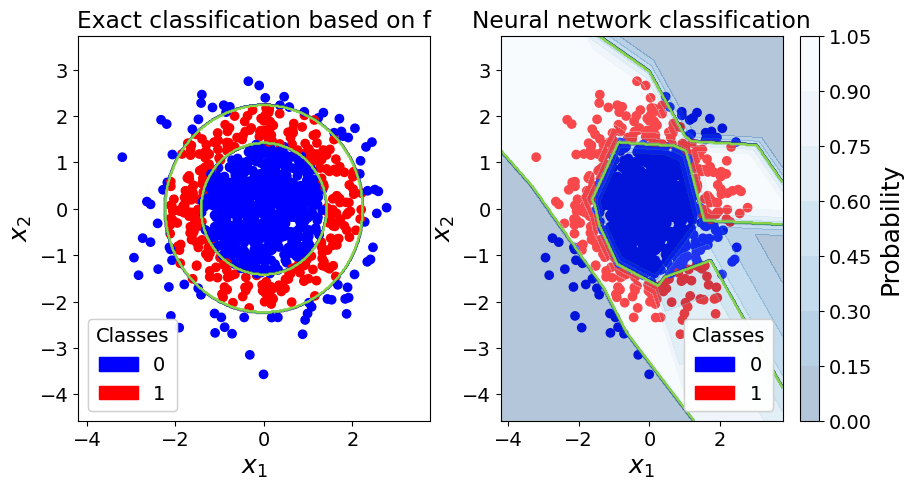} 
    \end{center}
        \caption{Training and testing on in-distribution data.}
        \label{fig:graph1}
    \end{subfigure}
    \\
    \begin{subfigure}{0.75\textwidth}
    \begin{center}
        \includegraphics[scale=0.25]{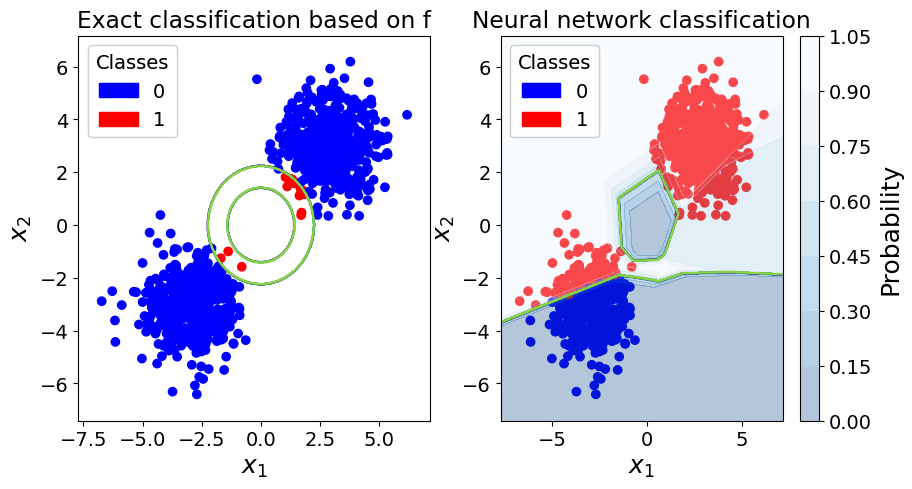}
    \end{center}
       \caption{Training on in-distribution data, and testing on out-of-distribution data.}
       \label{2dout}
    \end{subfigure}
    \\
    \begin{subfigure}{0.75\textwidth}
    \begin{center}
        \includegraphics[scale=0.25]{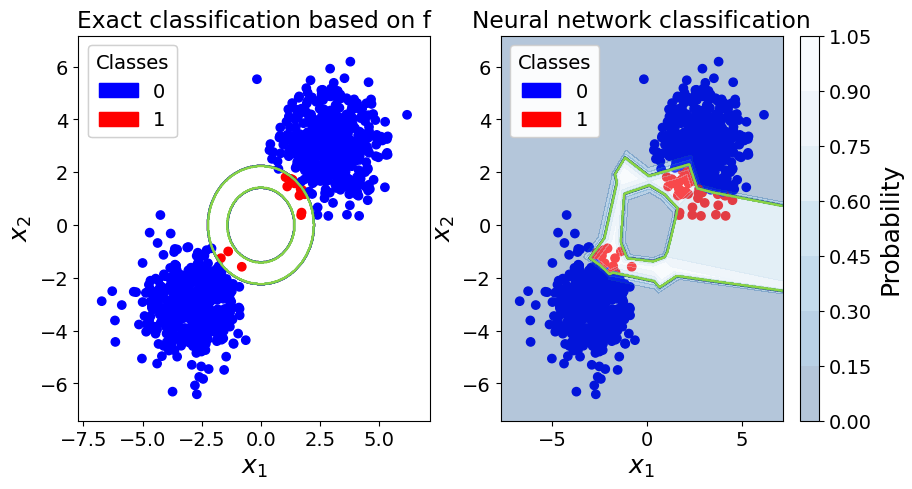}   
    \end{center}
        \caption{Training on in-distribution data, and testing on out-of-distribution data with outlier exposure}
        \label{fig:oetoy}
    \end{subfigure}
    \caption{\textbf{A simple math example to illustrate outlier exposure.} Three pairs of classification results in the toy example. From top to bottom: (a) is the result when training and testing data set is using the same normal distribution (in-distribution data). (b) is training on in-distribution data, and testing on out-of-distribution data. (c) Training on in-distribution data, and testing on out-of-distribution data with Outlier Exposure}
    \label{fig:toy}
\end{figure}

\end{exm}

In the real world, this relationship is not as trivial. This disparity of difficulty will cause the issue of out-of-distribution data to be more severe in real-world facial recognition applications.

In facial analysis applications, the out-of-distribution data could appear because of a different environment, such as lighting, or a different demographic or gender distribution during data collection. A face recognition algorithm transforms an image $\mathbf{x}$ into an output class $y$ indicating the identity of the person in that image. To train a face recognition model, a dataset needs to have many different images of faces. However, if the dataset uses photos taken indoors $P_{in}(\mathbf{x})$ to train the model, then images from outdoor lighting conditions $P_{out}(\mathbf{x})$ might cause the model to perform poorly. In another scenario, we want to train an algorithm to recognize personal attributes, such as emotion, from images of faces. People create a dataset of facial images with labels about emotion level sad to happy $P_{in}(y|\mathbf{x})$. However, cultural difference could result in different labels when a different group of people look at the same set of images $P_{out}(y|\mathbf{x})$. This might also result in wrong decisions from the model.

\subsection{Outlier Exposure with An Importance Re-weighting scheme}
\label{outlierexposure_section}

One approach we use to account for the out-of distribution data is by weighting the loss function in Equation~\eqref{outlier_exposure}. This is done by weighting the first term, the loss for in-distribution data. That is,
\begin{equation}
\begin{split}
   \E_{P_{in}(\mathbf{x}, y)}  L(f_{\theta}(\mathbf{x}),y) \approx -\frac{1}{n} \sum_{j=1}^n \sum_{i = 1}^K w_i \mathbf{1}[y_j = i] \log p_{\theta}(y_j = i| \mathbf{x}_j), 
\end{split}
\end{equation}
where the weights for each of the $I$ classes can be found by
\begin{align}
    w_i = \dfrac{n}{Iy_i^{\text{train}}} \label{weighting}
\end{align}
where $w_i$ is the weight used for the $i\text{th}$ class, $n$ is the total number of images in a dataset, and $y_i^{\text{train}}$ is the number of images in the $i\text{th}$ class, for $i \in [1, 2, \dots, I]$. Notice that since we consider binary classification, $I=2$, so we get:
\begin{align}
    w_1 &= \dfrac{n}{2y_1^{\text{train}}}, \\
    w_2 &= \dfrac{n}{2y_2^{\text{train}}}.
\end{align}
Additionally, we primarily use the rescaled weights, where $w_1 = 1$, and $w_2$ is scaled accordingly. This re-weighting scheme is chosen as it increases the importance of classifying the minority class.
 
This improves classification as more weight is placed on the positive class, which we consider to be the minority class of females for gender classification. Thus, when the loss function is being optimized, it will be penalized more heavily for errors made when training on the minority class.

In order to find the appropriate images for the outlier exposure data, we first need to quantify which images are outliers. We start by using the KL divergence to measure the difference between two probability distributions over a variable $x$. The two distributions we use are the pixel value distributions obtained from each dataset. The $x$-axis (pixels) are capped at $255$ due to the standard RGB scale. Every image in the dataset is a random variable $\mathbf{x} \in [0,1]^{n^2}$, where $n$ is the resolution value. The pixel average for the current image examined is defined as $\mathbf{x}_1$ and the overall distribution of pixel averages for all pixels is defined as $\mathbf{x}_2$, with the range for each of these averages between $0$ to $255$. The distance is defined as the distance between $\mathbf{x}_1$, $\mathbf{x}_2$ in the defined space. In this case we used the KL divergence which uses probability distributions. In probability distributions $P$ and $Q$ are defined by
\begin{equation}
    D_{\text{KL}}(P \parallel Q) = \sum_{x \in \mathcal{X}} P(x) \log\left(\frac{P(x)}{Q(x)}\right) \label{eq:kl_div}
\end{equation}
This divergence allows us to identify the 20\% of images from the UTKFace and FairFace dataset that represent the most significant outliers when compared to the distribution of the UTKFace dataset. These images are then compiled into separate folders to be used in Outlier Exposure testing where we analyze the distribution of each dataset based on the frequency of the pixel values. The pixel frequency distribution, which we used the KL divergence to find, represents the percentage of pixels with a given intensity featured in all images in a given dataset.

\begin{figure}[tph]
  \centering
  \includegraphics[width=0.75\textwidth]{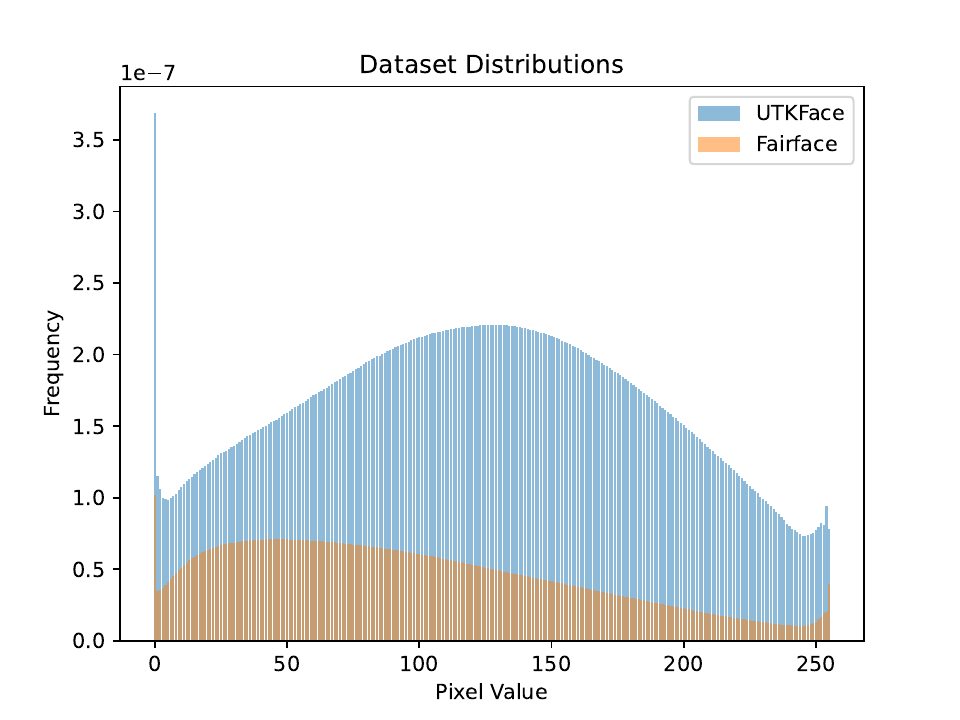}
  \caption{Distribution of Pixel Frequency in UTKFace $\&$ FairFace}
  \label{fig:pixels}
\end{figure}

 As can be seen in Figure~\ref{fig:pixels}, the mean RGB value of the UTKFace dataset is close to $120$, while the mean of the FairFace dataset is closer to $70$. Higher mean RBG values suggest that the dataset has brighter or more intense colors relative to lower mean RGB values \cite{ZabalaTravers2015}. From this, we deduce that the UTKFace dataset contains brighter images than the FairFace dataset, which might be due to image quality. That is an avenue for future research for classification models. 

We also find that the KL divergence between the two datasets to be $0.088$. Smaller KL divergences imply that two datasets are similar \cite{Galbraith2010}. This similarity is supported by Fig.~\ref{fig:pixels}, and may lead to issues when using this data in Outlier Exposure models because the two distributions are not at a level of distinction necessary for Outlier Exposure to be impactful \cite{hendrycks2018deep}.

However, there are major issues with comparing datasets using simple pixel values. Pixel values are incredibly sensitive to variation in lighting and are drastically affected by image transformations \cite{rs13112140}, computationally expensive \cite{Xing2019}, and do not consider important semantic content such as shapes and textures in an image \cite{Duan2021BridgingGB}. In addition, by using pixels we are considering every pixel value in an image, which includes background details that can lead to noisy and/or irrelevant features hindering accurate dataset comparisons \cite{NIPS2012_c399862d}. 

\begin{figure}
\centering
\begin{tikzpicture}[scale=2]
\node[] at (3.1,0){\includegraphics[width=.4\textwidth]{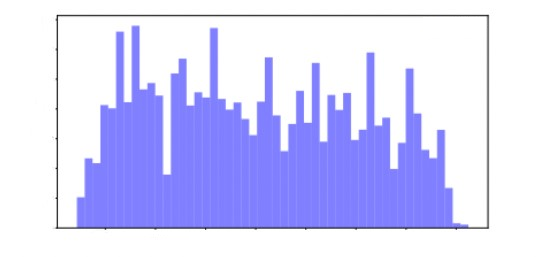}};
\node[label,rotate=90] at (1.5,0){Frequency};
\node[label] at (3.2,1){UTKFace};
\node[] at (5.9,0) {\includegraphics[width=.4\textwidth]{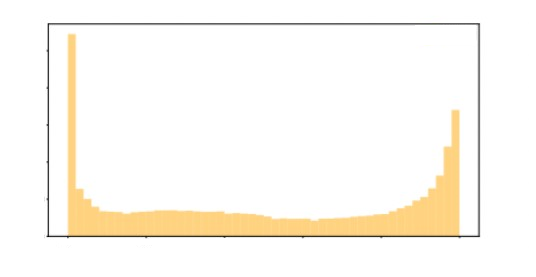}};
\node[label] at (5.9,1){Fairface};
\node[label] at (3.2,-0.75){Range: [0.018, 0.020]};
\node[label] at (5.9,-0.75){Range: [0, 1]};

\end{tikzpicture}
\caption{Frequency histogram of activation features in UTKFace and FairFace}
\label{fig:figure_label342347}
\end{figure}

Another method of comparing distributions can be done through activation features. Activation features, also known as feature maps, are the output maps of intermediate layers of a Convolutional Neural Network when the network performs forward propagation \cite{Nanni2022}. They have several benefits when compared to direct pixel values. Namely, they encode semantic information, reduce dimensionality to make computation more efficient, and are unaffected  by image transformations \cite{AZAM2023200233}. We found the histogram plots of the activation features for the UTKFace and FairFace datasets, shown in Fig.~\ref{fig:figure_label342347}.

The range of the features in the UTKFace dataset is limited as seen by the constraints of the $x$-axis. This range suggests that there is relatively minimal diversity in the contents of the UTKFace dataset. In addition, low activation values suggest that the feature is not abundantly apparent in the input data \cite{Horwath2020}. The dataset as a whole does not have an activation feature above $0.2$, suggesting that none of the features appear in a large number of the input images.

Comparatively, the FairFace dataset histogram in Fig.~\ref{fig:figure_label342347} has a remarkably different distribution. A large number of the features are in a similar range as the UTKFace dataset, but unlike UTKFace, some remain outside that range. This suggests that the distribution of the FairFace dataset may have a bigger spread than that of the UTKFace dataset. 

Additionally, this also confirms our belief that the two datasets are not nearly as similar as they appear when compared using pixel value frequencies. In order to quantify the difference between the two datasets we decided to use the KL divergece. This is important because the Outlier Exposure method of classification is only effective if two datasets have varied distributions.

The KL divergence for the two datasets when comparing activation features, instead of pixels, was over $2800$. A higher KL divergence values suggest that the two datasets are more dissimilar \cite{Galbraith2010}.

\subsection{Identifying and Quantifying Outlier Images and Datasets} \label{performance_metrics}
After confirming that the two datasets have significantly different distributions, we use this to modify the training of the neural network to classify out-of distribution data. One way of tackling the issue of classification of out-of distribution data is by attempting to shift the distribution of the two datasets together \cite{Weiss2016}. The hope is that by shifting the distributions closer, the model will better predict the test data as it is closer to the data seen before. 
To do this we first needed to identify the largest 20\% of  outliers in both the UTKFace and FairFace dataset; that is, we needed to find the images in each dataset that are farthest from the mean of the dataset. This would also allow us to quantify how far apart the datasets are from each other. This is relevant because if the two distributions are relatively similar, Outlier Exposure may not work \cite{hendrycks2018deep}. Furthermore, we can use the difference in distributions to modify the $\lambda$ in \eqref{outlier_exposure}. Although $\lambda$ was initially fixed to 0.5, we modified  it to be a function of the distribution differences of the training datasets. The equation for lambda was written as:
\begin{equation}
    \lambda(D_{\text{KL}}) := \tanh(D_{\text{KL}}), 
\end{equation}
where $D$ represents the KL pixel distance between the two distributions. We further modify this to change with respect to the number of epochs that have elapsed since training began in order to have the weight adjusted as the model sees more images. 
\begin{equation}
    \lambda(D_{\text{KL}},i) := \tanh(D_{\text{KL}})(1 - \cos(\frac{i \pi}{20}))
\end{equation}
where $i$ represents the current epoch from 1 to 20. This iteration causes $\lambda$ to increase with time, which places greater value on the second loss term during the last epochs of the model. In addition to changing $\lambda$, we also alter when we calculate the KL distance so as to yield different values. Initially, the KL distance of the two full distributions was implemented, then the KL distance of in-distribution and out-of-distribution in batch sizes of 16 images. This was done to see if smaller distributions might reveal intricacies within the dataset that were obscured at a more macroscopic level.

\subsection{Definition of Relevant Evaluation Metrics}
Here we define the metrics we used to evaluate our model. Due to the emphasis on the fairness of the model, we evaluated performance based on the confusion matrix and AUROC score. Through the use of the confusion matrix we find precision, recall, accuracy, and F1, with our goal being to maximize each metric, along with the AUROC score. Precision and recall can be calculated using the true positive, $TP$, false positive, $FP$, and false negative, $FN$, where
\begin{align}
    \text{Precision} &= \frac{TP}{TP+FP} \\
    \text{Recall} &= \frac{TP}{TP+FN}
\end{align}
Additionally, the F1 score is the harmonic mean between precision and recall. That is:
\begin{align}
    \text{F1} = \dfrac{2(\text{Precision})(\text{Recall})}{\text{Precision}+\text{Recall}}
\end{align}
The area under the ROC curve (AUROC) is a measure of performance, where a model with perfect predictions would have a score of $1$, and a model that was always incorrect would have a score of $0$. The ROC curves we obtain for each experiment can be found in Appendix~\ref{apdxroc}, along with the resulting confusion matrices for each experiment that we use to calculate the performance metrics in Appendix~\ref{apdxcm}.

\section{Experiments}
\label{sec:experiments}

\subsection{Datasets}
One dataset that we primarily use to train our model is the UTKFace dataset \cite{zhifei2017cvpr}. The dataset consists of over $20,000$ images labeled by age, gender, and race. Age is labeled numerically from $0-116$. For gender, $0$ is male, and $1$ is female. For race, $0$ is White, $1$ is Black, $2$ is Asian, $3$ is Indian, and $4$ is for any other races. 

We use the FairFace dataset \cite{kärkkäinen2019fairface} for training and testing, depending on the experiment. This dataset consists of around $100,000$ images, with labeling in a csv file consisting of labels for age range, gender, and race. To make our image names in the same format, we use a program to rename all the images in the FairFace folder to match the naming scheme of UTKFace. In addition, race is divided into seven categories, contrasting with UTKFace which has five. To account for this, we assigned White for Middle Eastern, and combined East Asian and Southeast Asian into Asian to mimic how UTKFace labels their images.

CIFAR-10 \cite{krizhevsky2009learning} is also used for the out-of-distribution dataset for Outlier Exposure. This dataset consists of $10$ classes of objects consisting of airplane, automobile, bird, cat, deer, dog, frog, horse, ship, and truck. This is chosen as an OOD set since we believe the pictures should appear different when compared to faces. 

Labeled Faces in the Wild (LFW) \cite{LFWTech} is used to test our model on a different dataset. LFW is used for gender classification. When looking at the gender distribution of this set, there are around $10,000$ male images, and around $3,000$ female images. It consists of $13,000$ images of different famous faces. This dataset is chosen as it is a widely used, large dataset consisting of faces.

CelebA \cite{liu2015faceattributes} is used similarly to the LFW dataset. The dataset consists of $202,599$ images of celebrities. Although they contain labels for many different traits, only the gender label is used. Unlike LFW, the number of male and female faces in the dataset is more balanced.

FairFace and UTKFace are also used experimentally as the OOD dataset. For each dataset, one approach is to use the entire dataset as the OOD set. Another approach is through using the outliers to create a smaller dataset consisting of the outliers of either FairFace of UTKFace.

\subsection{Dataset Processing}

In the pre-processing class, we make sure the labels between datasets are consistent, and we convert the image file path to an RGB array as an input for the neural network. Other research has previously shown that \cite{SHAO2017266} convolutional neural networks (CNN) perform better on RGB images relative to other forms of deep learning. As previous work showed \cite{10.1007/s11042-019-08373-8} neural networks take inputs of the same size and such that the images needed to have a fixed size before being used as inputs to the CNN. The image size we decided on was 32 pixels by 32 pixels, requiring resizing in order to avoid issues with computation time. Images that are modified in size struggle when processed by a neural network \cite{Hashemi2019}. To mitigate this problem we attempt to avoid minimizing images by a significant amount. The FairFace images are resized to 32 by 32 and all images are transformed into tensors necessary for CNN image processing. The tensors are also normalized before going through the network in order to make sure all the images are comparable.

\subsection{Network Infrastructure}

\begin{figure}[tph]
  \centering
  \includegraphics[width=0.8\textwidth]{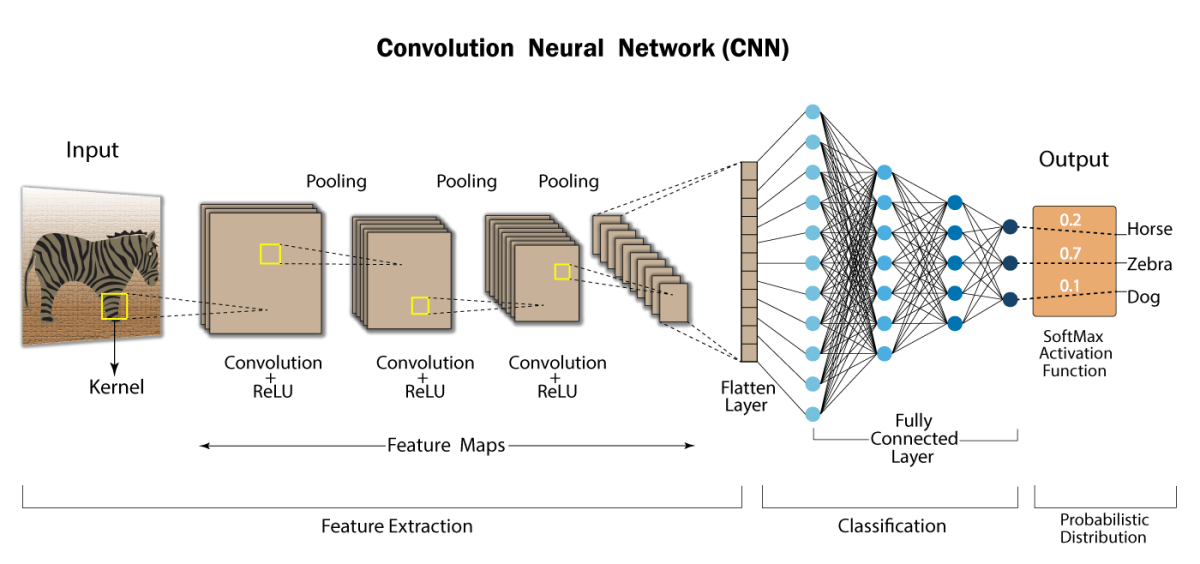}
  \caption{How a CNN Processes an Image.}
  \label{fig:figure_label3}
\end{figure}

In our model the network is a convolutional neural network (CNN) - a deep learning model for processing data that uses grid patterns \cite{Hubel1968} designed to learn the spatial hierarchies of different features in the image, making it optimal for facial recognition \cite{Fukushima1980}. The standard CNN contains three types of layers: convolution, pooling, and fully connected. The convolutional and pooling layers extract and process features through applying kernel matrices to preform convolutional and pooling operations. The fully connected layer maps it into an output that can be used for classification \cite{Yamashita2018}. For example, Fig~\ref{fig:figure_label3} shows how a convolutional network processes an inputted image.

Our first convolutional layer has an input channel of 3 RGB colors, 32 output channels and a kernel size of 5x5 - a common industry standard \cite{Hirahara2020}. Our second layer is a 2x2 kernel size pooling layer used to downsize the sample's spatial dimensions \cite{Nirthika2022} to make the CNN more effective. Average pooling is used (as opposed to max pooling) due to a more representative sample of the pixels in a region. Max pooling becomes skewed towards the dominant pixel features in an image region \cite{Zafar2022}. We perform a second convolutional layer to extract features from the pooled matrix before being concatenated to an $1$x$n$ vector and used as an input in the fully connected layers. The fully connected layers use randomly initialized weights to generate probabilistic interpretations of the random variable $x$, since pictures are loaded into the data loader at random \cite{Taye2023}. To extract information from each layer through weight functions the softmax activation function normalizes the output of the loss function so that back propagation can occur to update the randomly initialized weights. As the loss gets minimized the weights should converge to the desired function for how to interpret the input. This is done for any given number of epochs, where then the final softmax interpretation will be directly used to produce a label. The number of outputs matches the size of true-labels in classification. There is also a forward pass function in the class of our neural network that serves to push the image through the different layers of the network mentioned above. We use the $\tanh$ activation function over the ReLU activation function due to the non-linearity of tanh and its symmetry around zero \cite{DeRyck2021}. The function then returns the image after all the processing is complete.

We pretrain a reasonable convolutional neural network model by testing on in-distribution data. This is to make sure that we can explicitly address our proposed methods on improving the classification of out-of-distribution data.

When testing on in-distribution data, a model generally is able to have the highest accuracy. This is because when the two distributions of training and testing match, the model is well prepared for classifying those images.

\subsection{Training}

For training, every model is run with 20 epochs to compensate for large computation times. The optimizer we use was the ADAM optimizer \cite{kingma2017adam}. This optimizer was chosen over stochastic gradient descent because we found that it often converged more quickly while having the same loss values. 

\subsection{Testing on Out of Distribution Data}
To implement Outlier Exposure, it requires an in distribution and out-of-distribution dataset. We often use UTKFace as our in distribution dataset, which gets used as it would in a standard machine learning model. As seen in \eqref{OEIn}, the in distribution data passes through a standard loss function, which is typically cross entropy. The outlier exposure group is varied between different tests. As seen in \eqref{OEOut}, the outlier group does not pass through a typical loss function as the data for that group does not contain labels but only pictures. To account for this, we included a second loss function in the code \eqref{NewOEOut} which computes the cross entropy from a softmax distribution to a uniform distribution. This serves to minimize the program's confidence when predicting the out-of-distribution data, which should help prevent it from being overconfident when encountering other out-of-distribution data when testing. Potential future testing could involve a way for us to improve the weighting schemes by trying to mitigate the weight degeneracy in other ways. We are able to identify the weights of the two classes used in outlier exposure by using \eqref{Degeneracy} in order to make sure that our training set was not unduly biased towards male figures. 

In general, even if a model had been shown to be adept at recognizing images in the training distribution and accurately classifying them during testing, these results often do not generalize to out-of-distribution data. This means that we would train the model on one dataset, UTKFace for example, before testing on a different dataset - the goal being able to accurately detect out of distribution. It has been shown that this outcome is incredibly difficult for models since, although they may have a high confidence, their accuracy drops when tested on a different dataset's distribution \cite{hendrycks2017a}.

\begin{table}[tph]
    \centering
    \begin{tabular}{||c|c|c|c|c|c|c||}
    \hline
     \textit{Train Dataset} & \textit{Test Dataset} &\textit{Precision} & \textit{Recall} & \textit{Accuracy} & \textit{F1} & \textit{AUROC}\\
    \hline
     UTKFace & FairFace & 0.65 & 0.68 & 0.69 & 0.66 & 0.77 \\
    \hline
     FairFace & UTKFace & 0.87 & 0.86 & 0.87 & 0.86 & 0.95 \\
     \hline
     \end{tabular}
     \caption{Results from training and testing on OOD data based on gender. The model being used is tuned for in-distribution data, where all metrics are above 0.97 for UTKFace and 0.91 for FairFace.}
    \label{tab:ooddist}
\end{table}
As can be seen in Table~\ref{tab:ooddist}, the AUROC value of a model trained on UTKFace and tested on FairFace is substantially lower than when it is trained on Fairface and tested on UTKFace. This disparity is due to two major reasons. First, the FairFace dataset has more images which gives the model more data points to learn, which increases the accuracy of the model\cite{Hirahara2020}. Second, the FairFace dataset is also more balanced than the UTKFace data \cite{Wei2013} which leads to better accuracy. Those two changes make the FairFace dataset a better training dataset than the UTKFace dataset.

\label{sec:results}

\subsection{Improving Out-of-Distribution Classification}

 \begin{table}[tph]
     \centering
     \begin{tabular}{||c|c|c|c|c|c||}
    \hline
     \textit{Outlier Group} & \textit{Precision} & \textit{Recall} & \textit{Accuracy} & \textit{F1} & \textit{AUROC} \\
     \hline
     None & \textbf{0.65} & 0.68 & 0.69 & \textbf{0.66} & \textbf{0.77} \\
    \hline
     FairFace & 0.53 & 0.52 & 0.55 & 0.52 & 0.60 \\
    \hline
     UTKFace Outliers & 0.61 & \textbf{0.71} & \textbf{0.70} & \textbf{0.66} & 0.75 \\
    \hline
      FairFace Outliers & 0.58 & 0.66 & 0.66 & 0.62 & 0.70 \\
    \hline
       CIFAR-10 & 0.58 & \textbf{0.71} & 0.69 & 0.64 & 0.75 \\
     \hline
     \end{tabular}
     \caption{Results from Outlier Exposure when classifying based on gender, training on UTKFace and testing on Fairface}
     \label{tab:Outlier group choice}
 \end{table}

We found that the significance of the outlier group had a strong impact on the performance of the model. We believe UTKFace and UTKFace outliers did well due to the change in the distribution of UTKFace becoming more balanced, and potentially bring the distribution between the train and test set closer. However, as seen in Table~\ref{tab:Outlier group choice}, the original model without Outlier Exposure performs better. As seen in Figure~\ref{fig:pixels}, the distributions are already similar, providing an explanation to why Outlier Exposure did not work as well as we expected.

In attempt to take into account the differences in distributions between datasets, the samplings we use can be weighted. One approach we use is by taking the Outlier Exposure groups we created earlier from the 20\% with the greatest KL divergence, and adding that to the training data. This results in the model being exposed to a larger quantity of outliers to hopefully allow it to classify the images better during testing.

\begin{table}[tph]
    \centering
    \begin{tabular}{||c|c|c|c|c|c||}
        \hline
          \textit{Outlier Sample} & \textit{Precision} & \textit{Recall} & \textit{Accuracy} & \textit{F1} & \textit{AUROC} \\
         \hline
         None & 0.65 & 0.68 & 0.69 & 0.66 & 0.77 \\            
         \hline
          UTKFace Outliers & 0.70 & 0.68 & 0.70 & 0.69 & 0.77 \\
         \hline
          FairFace Outliers & \textbf{0.75} & \textbf{0.81} & \textbf{0.80} & \textbf{0.78} & \textbf{0.85} \\
         \hline
    \end{tabular}
    \caption{Results of increasing the samples for training while training on UTKFace and testing on FairFace}
    \label{tab:WeightSample}
\end{table}

As seen in Table~\ref{tab:WeightSample}, when the model is exposed to an increased sample size it performs much better than without.

Another approach to account for these differences can be by weighing the loss function to pay more attention to one class over another. This is useful in preventing a majority class from overwhelming the minority class. In particular, for gender we weight the female class higher than the male class. In order to observe this effect we did not implement outlier exposure. Our model is trained on UTKFace and tested on FairFace in order to observe the effect that the weighted sampling had independently on the model.

\begin{table}[tph]
    \centering
    \begin{tabular}{||c|c|c|c|c|c||}
        \hline
         \textit{Male Weight} & \textit{Female Weight} & \textit{Precision} & \textit{Recall} & \textit{F1} & \textit{AUROC} \\
         \hline
         1.0 & 1.0 & 0.57 & \textbf{0.65} & 0.61 & \textbf{0.70} \\
         \hline
         1.0 & 1.5 & \textbf{0.62} & 0.64 & \textbf{0.63} & \textbf{0.70} \\
         \hline
    \end{tabular}
    \caption{Results of modifying the weights of loss for male and female classes}
    \label{tab:Weightloss}
\end{table}

As seen in Table~\ref{tab:Weightloss}, we see that increasing the weights of the female class to 1.5 increases the precision while barely decreasing the recall. This means that overall, the model is classifying males and females with around the same accuracy as can be found in the corresponding confusion matrix.

\section{Main results}

We test our model on separate datasets, which simulates a scenario similar to a realistic application. We observe that by adding an outlier group, as outlined in Section~\ref{outlierexposure_section}, and weighting the sampling, as described in Section~\ref{weighting}, that the model performs better than without. We see that even when the extra outlier group is from a different sample entirely, the model is better at adapting to new situations.

 \begin{table}[tph]
 \centering
 \scalebox{0.85}{
     \begin{tabular}{||c|c|c|c|c|c||}
    \hline
     \textit{Training Group} & \textit{Precision} & \textit{Recall} & \textit{Accuracy} & \textit{F1} & \textit{AUROC} \\
     \hline
     FairFace & 0.46 $\pm$ 0.04 & 0.26 $\pm$ 0.01 & 0.60 $\pm$ 0.02 & 0.33 $\pm$ 0.01 & 0.57 $\pm$ 0.01 \\
    \hline
     FairFace w/ OE & \textbf{0.54 $\pm$ 0.01} & \textbf{0.30 $\pm$ 0.01} & \textbf{0.61 $\pm$ 0.02} & \textbf{0.38 $\pm$ 0.01} & \textbf{0.60 $\pm$ 0.02} \\
     \hline
     \end{tabular}}
     \caption{Averaged results from testing on LFW when classifying based on gender, in the form of mean $\pm$ standard error}
     \label{tab:Outlier group LFW}
     
 \end{table}

Since there is a strong imbalance between the male and female class in LFW, the loss is shown by Equation~\eqref{weighting}, using a male class weight of $1$ and female class weight of $1.3$. 

When finding the metrics, the female class was treated as the positive class, which in all prior datasets meant that measuring our accuracy metrics on the male or female class would yield similar values due to the balance between men and women in each set. However, in LFW there are around 10,000 males and 3,000 females, and as such, the metrics seem much lower as there is much less data for women. Regardless, as can be seen in Table~\ref{tab:Outlier group LFW}, our new model performs significantly better than the model that exclusively trains on FairFace. We take the average over five trials to observe any variance in both model. From these averages, it can be seen that there are improvements specifically in precision and recall, which shows that a greater percentage of women were accurately predicted.

When comparing the ROC curves for the two experiments, without outlier exposure the AUC is $0.58$. This is barely better than randomly guessing, as that would correspond to an AUC of $0.5$. Comparatively, with our model the AUC is $0.67$, which is a significant jump above the initial model.

An advantage of measuring the average over multiple trials is that the standard error can be found, which shows how much each model varied between tests. When comparing the values seen in Table~\ref{tab:Outlier group LFW}, the smallest standard error in each metric is split between the two models. However, the standard error for precision is much smaller in the model with outlier exposure when compared to the model without. 

\begin{table}[tph]
     \centering
      \scalebox{0.85}{
     \begin{tabular}{||c|c|c|c|c|c||}
    \hline
     \textit{Training Group} & \textit{Precision} & \textit{Recall} & \textit{Accuracy} & \textit{F1} & \textit{AUROC} \\
     \hline
     FairFace & 0.59 $\pm$ 0.020& 0.61 $\pm$ 0.007& 0.55 $\pm$ 0.007& 0.60 $\pm$ 0.009& 0.57$\pm$ 0.008 \\
    \hline
     FairFace w/ OE & \textbf{0.62 $\pm$ 0.007} & \textbf{0.64$\pm$0.007} & \textbf{0.58$\pm$0.008} & \textbf{0.63$\pm$0.005} & \textbf{0.59$\pm$0.008} \\
     \hline
     \end{tabular}}
     \caption{Averaged Results from testing on CelebA when classifying based on gender}
     \label{tab:Outlier group Celeba}
 \end{table}

Similarly to the LFW dataset, we set the weights for the male and female class to $1$ and $1.15$ respectively.
 
The model is also tested on the CelebA dataset to classify gender. The results shown in Table~\ref{tab:Outlier group Celeba} are consistent with the ones seen in Table~\ref{tab:Outlier group LFW}, where our improved model performs better in both fairness metrics and accuracy. 

Due to averaging multiple trials, we also observe the standard error between both models. When comparing the values seen in Table~\ref{tab:Outlier group Celeba}, it can be seen that on average, the standard error of the precision of the model with outlier exposure is smaller than that of the model without. A lower standard error means that the results are less varied and more consistent. 

Over the course of all experiments, we find that in general, our approach helps to reduce false negatives and false positives. This can be seen by observing the recall and precision in Table~\ref{tab:Outlier group LFW} and Table~\ref{tab:Outlier group Celeba}, in which both fairness metrics increase with our model. As such, this correlates to a lower false negative and false positive rate, which is very important as different applications suffer more from different false flagging. In contrast to the healthcare application mentioned prior, false positives can be particularly problematic in law enforcement applications, where a subject could be mislabeled as the culprit for a crime they did not commit. As such, being able to reduce both false positive and negative rates are important.

\section{Conclusions}
\label{sec:conclusions}

In this paper, we explore the concept of facial recognition classification. Specifically, we focus on how models struggle when confronted with data from different distributions during training and testing. Out-of-distribution data has been shown to significantly reduce the accuracy of facial recognition models. Throughout this paper, we document the results of a CNN model on different training and testing datasets. We also use KL distribution to identify outlier images from each dataset and incorporate Outlier Exposure into our model to see how it affected the model's accuracy and other metrics. We also attempt to weight the sampling of the different classes (male and female) and observe how the model's results change. Overall, we evaluate how facial recognition performs on out-of distribution data and conclude that outlier exposure can increase accuracy and other metrics of the model. We also conclude that utilizing weight sampling and outlier exposure can improve both the accuracy and the fairness of the model on out-of-distribution facial image sets. With the use of artificial intelligence and facial recognition in sectors such as law enforcement and healthcare, the lack of fairness and accurate classification of out-of-distribution data (often data $\&$ images of minority groups) has become an increasingly pressing issue. The methodologies we explore here can improve the metrics of these models, and we find that many techniques rely on having some knowledge about the dataset distributions, which requires a carefully considered implementation of these different methods to maximize the metrics of concern.

For this paper, all experiments are performed in Python using the Pytorch package, with data analysis performed using Scikit-learn package. The hardware we use is a 16 GB RAM and a 6 core, 2.6 GHz CPU. Additionally, our code will be made available upon request.

\subsection{Related \& Future Work}
While this paper is a good overview of current methods, there are alternate avenues that future researchers may want to explore to increase accuracy. One method is Geometric Sensitivity Decomposition, which works with feature norms of images after they have run through a neural network. Besides the importance weighting and outlier exposure perspectives we looked into, another region of focus is through bringing the distributions of the train and test datasets closer through the modification of the distributions. Working with this and incorporating geometric sensitivity decomposition would represent our next avenue to improve out-of-distribution images. 


Along the same lines of fairness transfer, \cite{schrouff2022diagnosing} develops a causal approach using conditional independence tests to characterize distribution shifts in healthcare machine learning, revealing that understanding such shifts can diagnose fairness discrepancies and suggesting potential mitigation strategies throughout the ML pipeline. Additional modifications to our CNN network could represent an area to improve the facial recognition of out-of-distribution images. 

\section*{Acknowledgments}
We would like to thank our mentor, Dr. Nicole Yang. This work was supported in part by the US NSF award DMS-2051019.


\clearpage
\appendix

\newpage

\section{ROC Curves for Experiments} 
\label{apdxroc}

\begin{figure}[tph]
\centering
\begin{tikzpicture}[scale=2]
\node[label,rotate=90] at (-0.25,0){True positive rate};
\node[] at (1.1,0){\includegraphics[width=.28\textwidth]{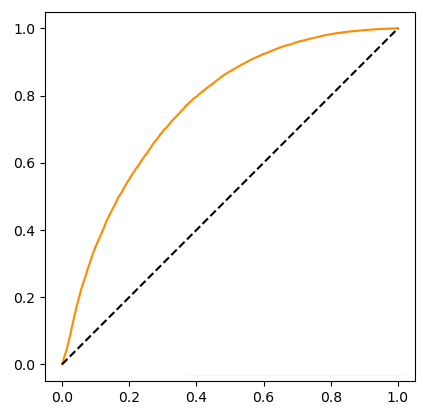}};
\node[label] at (1.1,1.2){UTKFace-FairFace-none};
\node[label] at (1.1,-1.15){False positive rate};

\node[] at (3.5,0) {\includegraphics[width=.3\textwidth]{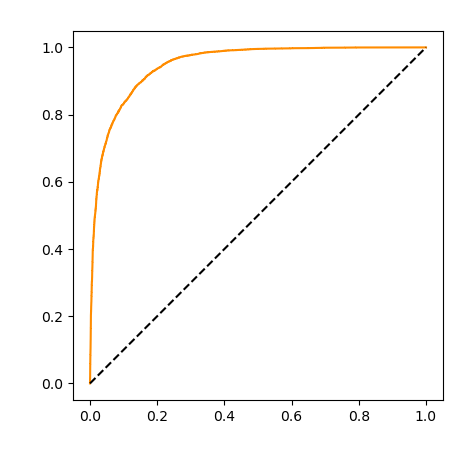}};
\node[label] at (3.6,1.2){FairFace-UTKFace-none};
\node[label] at (3.6,-1.15){False positive rate};

\node[] at (6,0) {\includegraphics[width=.28\textwidth]{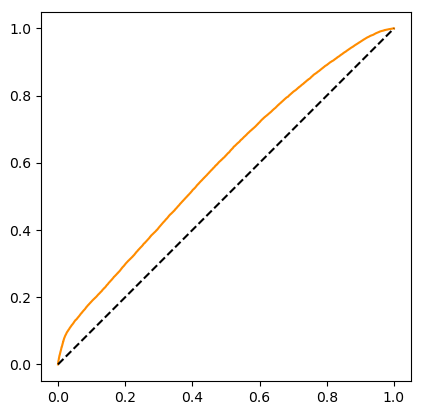}};
\node[scale=0.85, label] at (6,1.2){UTKFace-FairFace-FairFace};
\node[label] at (6.1,-1.15){False positive rate};
\end{tikzpicture}

\begin{tikzpicture}[scale=2]
\node[label,rotate=90] at (-0.25,0){True positive rate};
\node[] at (1.1,0){\includegraphics[width=.28\textwidth]{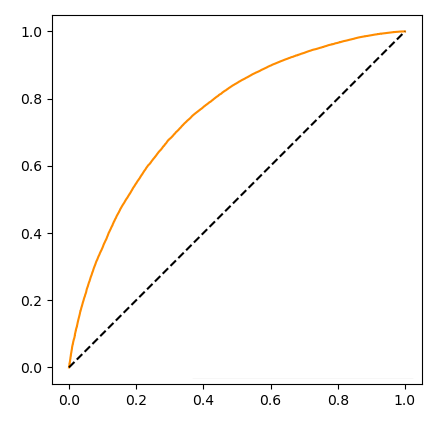}};
\node[scale=0.75,label] at (1.1,1.1){UTKFace-FairFace-UTKFace Outliers};
\node[label] at (1.1,-1.15){False positive rate};

\node[] at (3.5,0) {\includegraphics[width=.28\textwidth]{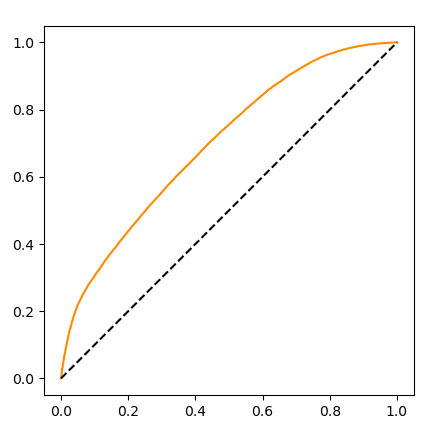}};
\node[scale=0.75,label] at (3.6,1.1){UTKFace-FairFace-Fairface Outliers};
\node[label] at (3.6,-1.15){False positive rate};

\node[] at (6,0) {\includegraphics[width=.28\textwidth]{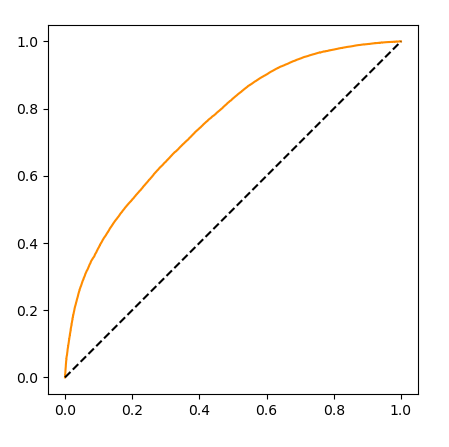}};
\node[scale=0.75,label] at (6,1.1){UTKFace-FairFace-CIFAR-10};
\node[label] at (6.1,-1.15){False positive rate};
\end{tikzpicture}

\begin{tikzpicture}[scale=2]
\node[label,rotate=90] at (-.2,-0.5){True positive rate};
\node[] at (1.1,-0.5){\includegraphics[width=.28\textwidth]{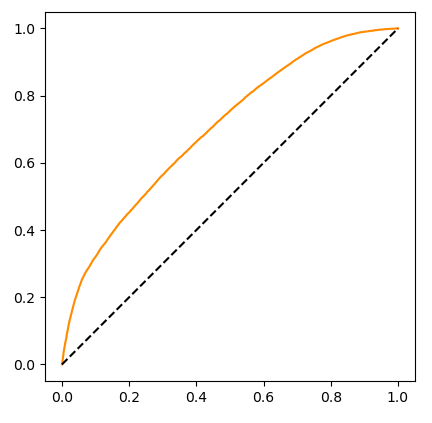}};
\node[scale=0.85, label] at (1,0.75){UTKFace-FairFace-none when loss weights are 1 for male, 1.5 for female};
\node[label] at (1.1,-1.65){False positive rate};
\end{tikzpicture}

\caption{ROC Curve from all experiments used, labelled in the order of ``train set-test set-outlier set". Orange is the classifier performance and the dashed line is for 50\% accuracy.}
\label{fig:figure_label32}
\end{figure}

\clearpage
\section{Confusion Matrix for Experiments} 
\label{apdxcm}

\begin{table}[tph]
     \centering
      \scalebox{0.85}{
     \begin{tabular}{||c|c|c|c|c||}
    \hline
     \textit{Train-Test-Outlier} & \textit{Male-Male} & \textit{Male-Female} & \textit{Female-Male} & \textit{Female-Female}   \\
     \hline
     UTKFace-FairFace & 33711 & 14274 & 12274 & 26484 \\
     \hline
     FairFace-UTKFace & 10818 & 1477 & 1573 & 9837 \\
     \hline
     UTKFace-UTKFace & 12079 & 226 & 312 & 11088 \\ \hline
     FairFace-FairFace & 42657 & 3754 & 3328 & 37004 \\
     \hline
     UTKFace-FairFace-UTKFace & 36306 & 15304 & 9679 & 25454 \\
     \hline
     UTKFace-FairFace-FairFace & 26010 & 19041 & 19975 & 21717 \\
     \hline
     UTKFace-FairFace-UTKFace Outliers & 35693 & 16079 & 10292 & 24679 \\
     \hline
     UTKFace-FairFace-FairFace Outliers & 33636 & 16952 & 12349 & 23806 \\
     \hline
     UTKFace-FairFace-CIFAR-10 & 36486 & 17144 & 9499 & 23614 \\
     \hline
     UTKFace-FairFace (weights 1-1.5) &32003 & 15503 & 13982 & 25255 \\
     \hline
     \end{tabular}}
     \caption{Confusion Matrices from all experiments used. Columns labeled by ``Predicted Label-Actual Label"}
 \end{table}

\bibliographystyle{siam}
\bibliography{main}

\end{document}